# AIR₅: Five Pillars of Artificial Intelligence Research

Yew-Soon Ong and Abhishek Gupta

*Abstract* – In this article, we provide an overview of what we consider to be some of the most pressing research questions currently facing the fields of artificial intelligence (AI) and computational intelligence (CI); with the latter focusing on algorithms that are inspired by various natural phenomena. We demarcate these questions using five unique *R*s – namely, (i) *rationalizability*, (ii) *resilience*, (iii) *reproducibility*, (iv) *realism*, and (v) *responsibility*. Notably, just as *air* serves as the basic element of biological life, the term AIR₅ – cumulatively referring to the five aforementioned *R*s – is introduced herein to mark some of the basic elements of artificial life (supporting the sustained growth of AI and CI). A brief summary of each of the *R*s is presented, highlighting their relevance as pillars of future research in this arena.

## I. INTRODUCTION

The original inspiration of *artificial intelligence* (AI) was to build autonomous systems capable of demonstrating human-like intelligence. Likewise, the related field of *computational intelligence* (CI) emerged in an attempt to artificially recreate the consummate learning and problem-solving ability depicted by various natural phenomena – including the workings of the biological brain. However, in the present-day, the combined effects of (i) the relatively easy access to massive and growing volumes of data, (ii) rapid increase in computational power, and (iii) the steady improvements in data-driven *machine learning* (ML) algorithms [1, 2], have played a major role in helping modern AI systems vastly surpass humanly achievable performance across a variety of applications. In this regard, some of the most prominent success stories are IBM's Watson winning Jeopardy! [3], Google DeepMind's AlphaGo program beating the world's leading Go player [4], their AlphaZero algorithm learning entirely via "self-play" to defeat a world champion program in the game of chess [5], and Carnegie Mellon University's AI defeating four of the world's best professional poker players [6].

With the rapid development of AI technologies witnessed over the past decade, there is general consensus that the field is indeed primed to have a significant impact on society as a whole. Given that much of what has been achieved by mankind is a product of human intelligence, it is amply clear that the possibility of augmenting cognitive capabilities with AI holds immense potential for realizing novel breakthroughs in critical areas such as healthcare, renewable energy, economics, etc. That said, there continue to exist key scientific challenges that require foremost attention for the concept of AI to be more widely trusted, accepted, and integrated within the fabric of society. In this article, we demarcate these challenges using five unique *R*s – viz., (i) $R_1$: *rationalizability*, (ii) $R_2$: *resilience*, (iii) $R_3$: *reproducibility*, (iv) $R_4$: *realism*, and (v) $R_5$: *responsibility* – which, in our opinion, shall form five pillars of future AI research. In summary, just as *air* serves as the basic element of biological life, the term AIR₅ – cumulatively referring to the five *R*s – is introduced herein to mark some of the basic elements of artificial life.

With this, the remainder of the article is organized to provide a brief summary of each of the aforementioned *R*s, highlighting their fundamental relevance towards the sustained growth of AI in the years to come.

## II. $R_1$: RATIONALIZABILITY OF AI SYSTEMS

Currently, many (if not most) of the innovations in AI are driven by ML techniques centered around the use of so-called *deep neural network* (DNN) models [2]. The design of DNNs is loosely based on the complex biological neural network that makes up a human brain – which (unsurprisingly) has drawn significant interest among CI researchers as a dominant source of intelligence in the natural world. However, DNNs are often criticized for being highly *opaque*. It has been widely acknowledged that although these models can frequently attain remarkable prediction accuracies, their layered non-linear structure makes it exceeding difficult (if not impossible) to unambiguously *interpret* why a certain set of inputs leads to a particular output / prediction / decision. As a result, at least at present, these models have come to be viewed mainly as *black-boxes* [7, 8].

With the above in mind, it is argued that for humans to cultivate greater acceptance of modern AI systems, their workings (and their consequent outputs) need to be made more *rationalizable – i.e., possess the ability to be rationalized (interpreted / logically explained)*. Moreover, the need for rationalizability cannot be compromised in safety critical applications, such as medical diagnosis, self-driving cars, etc., where peoples' lives are immediately at stake. Incidentally, a well-known study revealing the threat of opacity in *neural networks* (NNs) is the prediction of patient mortality in the area of community-acquired pneumonia [9]. While NNs were *seemingly* the most accurate for this task, an alternate (less accurate but more interpretable) rule-based system was later found to learn the following rule from one of the pneumonia datasets: $HasAsthma(x) \Rightarrow LowerRiskOfDeath(x)$ [10]. Even though the inferred rule is patently dubious, it reflects a definite (albeit misleading) pattern in the data (used to train the system) that may also hamper a NN. Unfortunately, the inability to rigorously examine the NN in such delicate situations often tends to preclude its practical applicability; as was the case for the patient mortality prediction problem. In fact, similar situations are likely to be encountered in general

Yew-Soon Ong and Abhishek Gupta are with the Data Science and Artificial Intelligence Research Centre (DSAIR), School of Computer Science and Engineering, Nanyang Technological University (NTU), Singapore. E-mail: {asysong, abhishekg}@ntu.edu.sg

scientific / engineering disciplines as well, where an AI system is usually required to be consistent with certain fundamental physical laws. Therein, the availability of rationalizable models, that are grounded in established theories, can go a long way in protecting against the learning of spurious patterns from raw data [11].

On a different note, it is contended that while the scope for interpretability can provide insights into the reasoning behind a model's prediction / decision, it does not reflect its *level of confidence*. To elaborate, given a previously unseen input data point (especially one that is outside the regime of the dataset used for model training), it only makes sense for a rational ML model to be unsure about its prediction. *Clearly representing the degree of uncertainty through principled predictive distributions is therefore crucial to prevent misleading the end-user.* For this reason, *uncertainty quantification* is considered to be another important facet of AI rationalizability. In this regard, it turns out that although DNNs are (rightly) considered to be state-of-the-art among ML techniques, they do not (as of now) satisfactorily represent uncertainties [12]. This sets the stage for future research endeavors in probabilistic AI and ML, with some recent foundational works in this arena being presented in [13, 14].

### III. $R_2$: RESILIENCE OF AI SYSTEMS

Despite the impressive progress made in AI, latest research has shown that even the most advanced models (e.g., DNNs) have a peculiar tendency of being easily fooled [15]. Well-known examples of this weakness have been put forth in the field of computer vision [16], where the output of a trained DNN classifier is found to be drastically altered by simply adding a small, well-tuned additive perturbation to an input image. Generally, the added perturbation (also known as an *adversarial attack*) is so small that it is completely imperceptible to the human eye; and yet causes the DNN to misclassify. In extreme cases, attacking only a single pixel of an image may also suffice in fooling various types of DNNs [17]. A particularly instructive illustration of the overall phenomenon is described in [18], where, by adding a few black and white stickers to a "Stop" sign, an image recognition AI was fooled into classifying it as a "Speed Limit 45" sign. It is worth highlighting that similar results have been reported in speech recognition applications as well [19].

While the consequences of such gross misclassification can evidently be dire, the aforementioned ("Stop" sign) case-study is especially alarming for industries like that of self-driving cars. For this reason, there have been targeted efforts over recent years towards attempting to make DNNs more *resilient – i.e., possess the ability to retain high predictive accuracy even in the face of adversarial attacks (input perturbations)*. To this end, some of the proposed approaches include brute-force adversarial training [20], gradient masking / obfuscation [21], defensive distillation [22], and network add-ons [23], to name a few. Nevertheless, the core issues are far from being eradicated, and demand significant future research attention [24].

Incidentally, in contrast to adversarial attacks that are designed to occur after a fully trained model is deployed for operation, *data poisoning* has emerged as a different kind of attack that can directly cripple the training phase. Specifically, the goal of an attacker in this setting is to *subtly* adulterate a training dataset (either by adding new data points [25] or modifying existing ones [26]), such that the learner is forced to learn a bad model. Clearly, ensuring resilience against such attacks is of paramount importance, as the main ingredient of all ML systems – namely, the training data itself – is drawn from the outside world (for e.g., in the form of online product reviews) wherein it is highly exposed to being intentionally or unintentionally attacked / corrupted [27].

### IV. $R_3$: REPRODUCIBILITY OF AI SYSTEMS

A common challenge faced while building DNNs, and ML models in general, is the replication crisis [28]. Basically, many of the key results reported in the literature are found to be difficult to reproduce by others. As noted in [29], for any claim to be believable and informative, reproducibility is a minimum necessary condition. Thus, ensuring performance reproducibility of AI systems is vital for maintaining their trustworthiness. In what follows, we briefly discuss two complementary tracks in pursuit of the desired outcome.

The first obstacle in the path of achieving reproducibility is the large number of *hyperparameters* (e.g., neural architectural choices) that need to be configured prior to training a model on any given dataset [30]. Even though these configurations typically receive secondary treatment among the core ingredients of a model or learning algorithm, their precise setting can considerably affect the efficacy of the learning process. Consequently, the lack of expertise in hyperparameter selection can sometimes lead to unsatisfactory behavior of the resultant learned model. Said differently, the model fails to live up to its true potential – as may have been reported in a scientific publication. With the above in mind, a promising alternative to manual configurations is to *automate* the process of hyperparameter selection, by formulating it as a global optimization problem. To this end, a range of algorithms, encompassing evolutionary strategies [31] as well as Bayesian optimization techniques [32], have been proposed in recent years, making it possible to select near-optimal hyperparameters without the need for any human intervention. The general approach thus falls under the scope of *AutoML* (automated machine learning) [33].

Notably, one of the cutting-edge topics of interest in AutoML is the capacity for automatic *reuse* of knowledge across distinct (but *related*) problems or datasets; which is made feasible under the novel concept of *transfer / multi-task learning in optimization* [34-36]. An associated research strand in the realm of CI is that of *memetic computation* – where *a meme takes the form of a basic unit of computationally encoded problem-solving knowledge that can be learned from one task and transmitted to another* [37]. As illustrated by ongoing research efforts [38], the main motivation behind the exploitation of recurring (shared) building-blocks of learned knowledge is the promise of cross-domain performance generalization in AI.

Alongside optimum hyperparameter selection, the second track for realizing AI reproducibility is the simple practice of openly sharing source codes and datasets corresponding to published scientific papers. While this practice is indeed followed by many, a recent survey suggests that the current documentation practices at top AI conferences render the reported results mostly irreproducible [39]. Thus, certain (universally agreed) software standards, pertaining to code documentation, data formatting, etc., are urgently needed for rigorous validation studies to be carried out easily.

## V. $R_4$: REALISM OF AI SYSTEMS

The three *R*s presented so far mainly focus on the predictive performance of AI systems. In this section, we shift our attention to a different feature that, looking ahead, is deemed vital for the seamless assimilation of AI into society.

In addition to absorbing vast quantities of data to support complex decision-making, AI has shown promise in domains involving intimate human interactions as well. Some prominent examples include the everyday usage of smart speakers (like Google Home devices or Amazon Alexa), the improvement of education through virtual tutors [40], and providing psychological support to Syrian refugees through the use of chat-bots [41]. *A common facet among all the aforementioned applications is the need for relatability. In other words, applications such as the ones above shed light on the need to introduce a human element into AI; which could potentially be achieved by enhancing its proficiency in recognizing, interpreting, and expressing real-life emotions and sentiments.* Various research threads have emerged in pursuit of such *realism* in autonomous intelligent systems, encompassing topics like *affective computing* [42] and *collective intelligence* for humanizing AI [41].

On one side, the key challenge facing the field of affective computing is the development of systems that can detect and process *multimodal data streams*. The motivating rationale stems from the observation that different people express themselves in different ways, utilizing diverse modes of communication (such as speech, body-language, facial expressions, etc.) to varying extent. Therefore, in most cases, the fusion of visual and aural information cues is able to offer a more holistic understanding of a person's emotion; at least in comparison to the best *unimodal* analysis techniques that process separate emotional cues in isolation [43, 44].

In contrast to affective computing, which is focused on a specific class of learning problems, collective intelligence is a meta-concept (incorporating any underlying learning problem) that deals with explicitly tapping on the wisdom of a "crowd of people" to guide AI. For instance, it was reported in [45] that through a *crowdsourcing* approach to selecting relevant features in big datasets, near state-of-the-art performance could be reached within a short period of time. Nonetheless, over and above the obtained predictive accuracy, it is contended that introducing a human element into an otherwise mechanized procedure of learning from raw data tends to have the more significant effect of enhancing the legitimacy and acceptability of AI in society's eye.

## VI. $R_5$: RESPONSIBILITY OF AI SYSTEMS

Last but certainly not least, we refer to the IEEE guidelines on Ethically Aligned Design which states the following [46]:

*"As the use and impact of autonomous and intelligent systems become pervasive, we need to establish societal and policy guidelines in order for such systems to remain human-centric, serving humanity's values and ethical principles."*

Thus, it is this goal of *building ethics into AI* [47] that we subsume under the final *R*, namely, *responsibility*; with the term "ethics" being defined in [48] as *a normative practical philosophical discipline of how one should act towards others*.

As has previously been mentioned, perhaps the most astonishing outcome of modern (black-box) AI technologies is their ability to uncover and learn from complex patterns buried in vast volumes of data, gradually attaining performance levels that far exceed human limits. However, not so surprisingly, it is their remarkable strength that has also turned out to be a matter of grave unease; with dystopian scenarios of robots taking over the world being frequently discussed nowadays [49]. Accordingly, taking inspiration from the fictional organizing principles of Isaac Asimov's robotic-based world, the present-day AI research community has begun to realize that machine ethics play a central role with regards the manner in which autonomous systems are permitted to interact with humans and with each other [50].

That said, clearly demarcating what constitutes ethical machine behavior, such that precise laws can be created around it, is not a straightforward affair. While existing frameworks have largely placed the burden of codifying ethics on AI developers, it was noted in [51] that ethical issues pertaining to intelligent systems are beyond the grasp of the original system developers. Indeed, there exist several subtle questions spanning matters of privacy, public policy, national security, among others, that demand a joint dialogue between computer scientists, legal experts, political scientists, and ethicists [52]. For example, a collection of illustrative questions that will likely be raised in the imminent future, but are difficult to objectively resolve, are listed below.
(i) In terms of privacy, to what extent should AI systems be allowed to probe and access one's data from surveillance cameras, phone lines, or emails?
(ii) How should policies be framed for autonomous vehicles to trade-off a small probability of human injury against near certainty of major material loss to private or public property?
(iii) In national security (defense) applications, how should autonomous weapons comply with humanitarian law while simultaneously preserving their original design objectives?
It is not hard to imagine the difficulty of arriving at a consensus when dealing with issues of the aforementioned type. The challenge is further exacerbated by the fact that ethical correctness is often subjective, and can vary across societies and individuals. Hence, the goal of building ethics into AI is unquestionably a matter of vital future importance that demands substantial research investment.

In conclusion, it is worth highlighting that the capabilities discussed in $R_1$ (i.e., rationalizability of AI systems) also form key ingredients towards attaining greater responsibility in AI, making it possible for an autonomous intelligent system to explain its actions under the framework of human ethics. In fact, the ability to do so is necessitated by a "right to explanation", as is implied under the European Union's General Data Protection Regulation [53].